\documentclass[conference]{IEEEtran}
\IEEEoverridecommandlockouts

\usepackage{cite}
\usepackage{amsmath,amssymb,amsfonts}
\usepackage{algorithmic}
\usepackage{graphicx}
\usepackage{textcomp}
\usepackage{hyperref}
\usepackage{xcolor}
\usepackage{multirow}
\usepackage{xcolor}
\usepackage{booktabs} 
\usepackage{adjustbox}

\newcommand\ourmethod{WhisperNER}

\def\BibTeX{{\rm B\kern-.05em{\sc i\kern-.025em b}\kern-.08em
    T\kern-.1667em\lower.7ex\hbox{E}\kern-.125emX}}
\begin{document}

\title{WhisperNER: Unified Open Named Entity and Speech Recognition}

\author{\IEEEauthorblockN{Gil Ayache}
\IEEEauthorblockA{\textit{aiOla Research} \\
\texttt{gil@aiola.com}}
\and
\IEEEauthorblockN{Menachem Pirchi$^\dagger$ \thanks{$^\dagger$Work done while at aiOla Research.}}
\IEEEauthorblockA{\textit{Independent Researcher} \\
}
\and
\IEEEauthorblockN{Aviv Navon}
\IEEEauthorblockA{\textit{aiOla Research} \\
}
\and
\IEEEauthorblockN{Aviv Shamsian}
\IEEEauthorblockA{\textit{aiOla Research} \\
}
\and
\IEEEauthorblockN{Gill Hetz}
\IEEEauthorblockA{\textit{aiOla Research} \\
}
\and
\IEEEauthorblockN{Joseph Keshet}
\IEEEauthorblockA{\textit{aiOla Research} and \textit{Technion}}
}

\maketitle

\begin{abstract}
Integrating named entity recognition (NER) with automatic speech recognition (ASR) can significantly enhance transcription accuracy and enrich its content. We introduce WhisperNER, a novel model that facilitates joint speech transcription and entity recognition. WhisperNER supports open-type NER, enabling recognition of various entities during inference. Building on recent advancements in open NER research, we augment a large synthetic dataset with synthetic speech samples. This approach enables us to train WhisperNER on numerous examples with various NER tags. During training, the model is prompted with NER labels and optimized to produce the transcribed utterance alongside the corresponding tagged entities. For evaluation, we generate synthetic speech for commonly used NER benchmarks and annotate existing ASR datasets with open NER tags. Our experiments show that WhisperNER outperforms natural baselines in both out-of-domain open-type NER and supervised fine-tuning.
\end{abstract}


\section{Introduction}
\label{sec:intro}

Recent advancements in both speech processing and natural language processing (NLP) have dramatically expanded the capabilities of automatic speech recognition (ASR) systems. These improvements, driven by large-scale models like Whisper~\cite{radford2023robust}, have significantly enhanced transcription accuracy. Meanwhile, the emergence of powerful NLP models, particularly large language models (LLMs), has led to major breakthroughs in tasks like language understanding, text generation, and named entity recognition (NER).
Traditionally, speech-driven applications have relied on pipeline architectures, where ASR systems transcribe speech into text, which is then processed by NLP models for tasks such as sentiment analysis, question answering, and NER. While this approach has proven effective across a range of applications, it suffers from error accumulation --- where transcription errors in the ASR stage propagate through the pipeline, reducing the performance of downstream NLP tasks. This challenge is particularly evident in complex tasks requiring high accuracy, such as NER.

A growing area of interest is the integration of speech models into a variety of NLP tasks beyond transcription, 
allowing models to process spoken language directly and eliminate intermediate
stages~\cite{bastianelli2020slurp,shon2022slue1,shon2022slue2,arora2024evaluation,arora2024universlu}. For instance, spoken language understanding (SLU) models have been developed to perform tasks such as intent detection and NER~\cite{yang2023chinese,meeus2023whisper,li2024prompting} directly from speech, bypassing the need for separate ASR components. These end-to-end models have shown promising results by jointly optimizing speech and language understanding.

\begin{figure}[t]
    \centering
    \includegraphics[width=1.\linewidth]{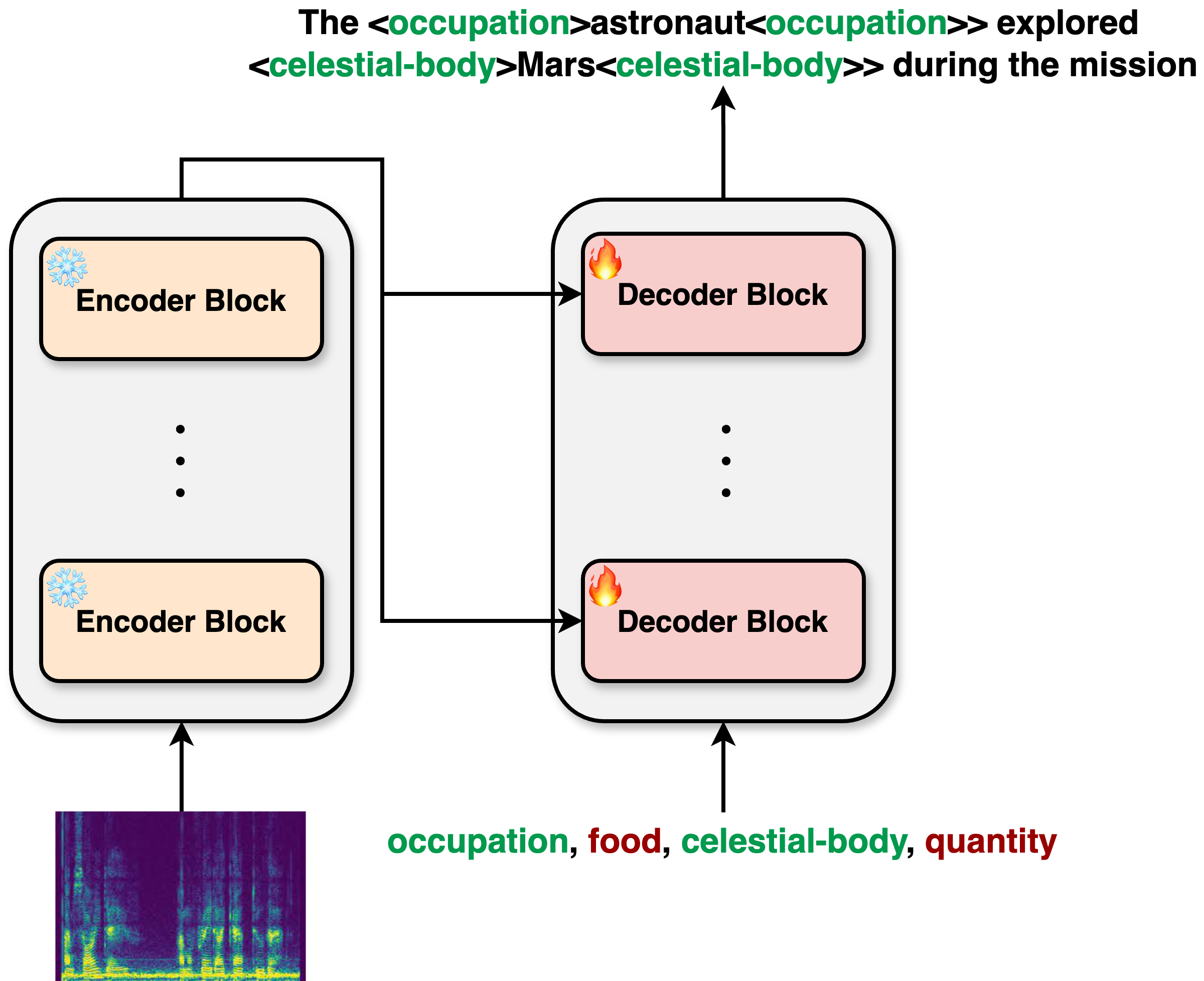}
    \caption{The architecture of \textit{WhisperNER}. A set of entity types is provided as a prompt to the decoder. During training, we provide positive (green) and negative (red) entities. At inference time,  the model can generalize to novel entity types not observed during training.}
    \label{fig:arch}
\end{figure}

\begin{table*}[t]
\small
\centering
\caption{Zero-shot, out-of-domain evaluation on three open-type NER speech benchmarks. The Params column indicates the extra parameters of baseline methods w.r.t \ourmethod{} (since baselines use a pipeline that consists of the Whisper model and the NLP model, while \ourmethod{} uses only the Whisper architecture). We bold the best results and underline the second-best results.}
\begin{adjustbox}{max width=1.\textwidth}
\begin{tabular}{lcccccccccccccc}
\toprule
 & \multicolumn{2}{c}{VoxPopuli-NER} && \multicolumn{2}{c}{LibriSpeech-NER} && \multicolumn{2}{c}{Fleurs-NER}  && \multicolumn{2}{c}{Average} \\ 
 \cmidrule{2-3} \cmidrule{5-6} \cmidrule{8-9} \cmidrule{11-12}
& F1 $\uparrow$ 
& WER $\downarrow$  
&& F1 $\uparrow$ 
& WER $\downarrow$
&& F1 $\uparrow$ 
& WER $\downarrow$
&& F1 $\uparrow$ 
& WER $\downarrow$
&& Params $\downarrow$ \\
 \midrule 
GNER-T5-base & $45.31$ &  $\mathbf{8.32}$ && $35.68$ & $\mathbf{5.51}$ && $47.15$ & $\underline{8.52}$  && $42.72$ & $\mathbf{7.45}$ && $+248$M \\
 NuNER & $52.18$ &  $\mathbf{8.32}$ && $43.65$ & $\mathbf{5.51}$ && $53.11$ & $\underline{8.52}$  && $49.64$ & $\mathbf{7.45}$ && $+459$M \\ 
 GLiNER &  $53.55$ &  $\mathbf{8.32}$  && $48.97$ & $\mathbf{5.51}$ && $\underline{54.35}$ & $\underline{8.52}$  && $52.29$ & $\mathbf{7.45}$ && $+459$M \\ 

 \midrule
 \ourmethod{}-BIO & $\underline{55.79}$ & $9.51$ && $\underline{50.40}$ & $5.95$ && $\mathbf{54.42}$ & $\mathbf{8.22}$ && $\mathbf{53.53}$ & $\underline{7.89}$ && $-$ \\
 \ourmethod{} & $\mathbf{56.25}$ & $\underline{9.22}$ && $\mathbf{50.84}$ & $\underline{5.82}$ && $53.50$ & $10.26$ && $\mathbf{53.53}$ & $8.44$ && $-$ \\
 \bottomrule
\end{tabular}
\end{adjustbox}
\label{tab:zero_shot_open}
\end{table*}
Recently, open NER~\cite{zhou2023universalner,zaratiana2023gliner, bogdanov2024nuner} has gained significant attention for its ability to generalize to new and unseen entities, offering a flexible alternative to traditional NER systems that are restricted by fixed entity types. Zhou \emph{et al.} (2023) \cite{zhou2023universalner} introduce a targeted distillation method, where student models are trained through instruction tuning. Zaratiana \emph{et al.} (2023) \cite{zaratiana2023gliner} and Bogdanov \emph{et al.} (2024) \cite{bogdanov2024nuner} suggested using a tailored, small size transformer encoders, designed specifically for the NER problem, and showed that these models can outperform LLM based NER models. Sainz \emph{et al.} (2023) \cite{sainz2023gollie} propose a method for improving information extraction results through annotation
guidelines, and Ding \emph{et al.} (2024) \cite{ding2024rethinking} explored the importance of ``negative instances,'' i.e, non-entity spans, for improving contextual information and label boundaries. However, integrating open type NER directly into ASR systems remains an unexplored area, as current end-to-end NER models can only operate with a fixed predefined entity set.

In this paper, we propose \ourmethod{}, a unified model that simultaneously performs speech transcription and named entity recognition, based on the Whisper ASR model~\cite{radford2023robust}. By integrating NER directly into the ASR process, \ourmethod{} eliminates the need for separate ASR and NER components, effectively reducing the risk of error propagation. 
\ourmethod{} closes the gap between SLU based NER approaches and the recent NLP-based open-type NER methods. 
By supporting open-type NER, \ourmethod{} can identify a broad range of entities at inference, making entity extraction from speech more accurate and flexible.
To build \ourmethod{}, we leverage a large-scale synthetic dataset released by Bogdanov \emph{et al.} (2024) ~\cite{bogdanov2024nuner}. We augment a subset of the textual open NER dataset with corresponding synthetic audio samples. The final dataset contains $350$K samples and $1.8$M unique entity types. Our approach draws motivation from recent approaches, which incorporate prior information and context into the decoding process by prompting the model on domain information~\cite{liao2023zero} or keyword phrases~\cite{shamsian2024keyword,Li2023AMT}. We train \ourmethod{} by prompting the model with NER type, for the task of joint transcription and NER tagging as depicted in Figure~\ref{fig:arch}. At inference time, \ourmethod{} can generalize to novel entity types not observed during training.

This paper makes the following contributions: (i) propose \ourmethod{}, a novel model for joint open NER and ASR; (ii) introducing new speech and text NER and open-NER benchmarks; (iii) conduct extensive experiments to demonstrate the superiority of \ourmethod{} over existing methods.
To encourage future research and reproducibility, we make our source code, datasets, and models publicly available.


\section{Our Method}
\label{sec:method}

In this section, we describe our approach, \ourmethod{}, which extends the Whisper ASR model~\cite{radford2023robust} to allow for joint open NER and speech recognition. The model is designed to output a tagged transcription sequence given an audio input, effectively performing both tasks simultaneously.

\begin{figure}[t]
    \centering
    \includegraphics[width=1.\linewidth]{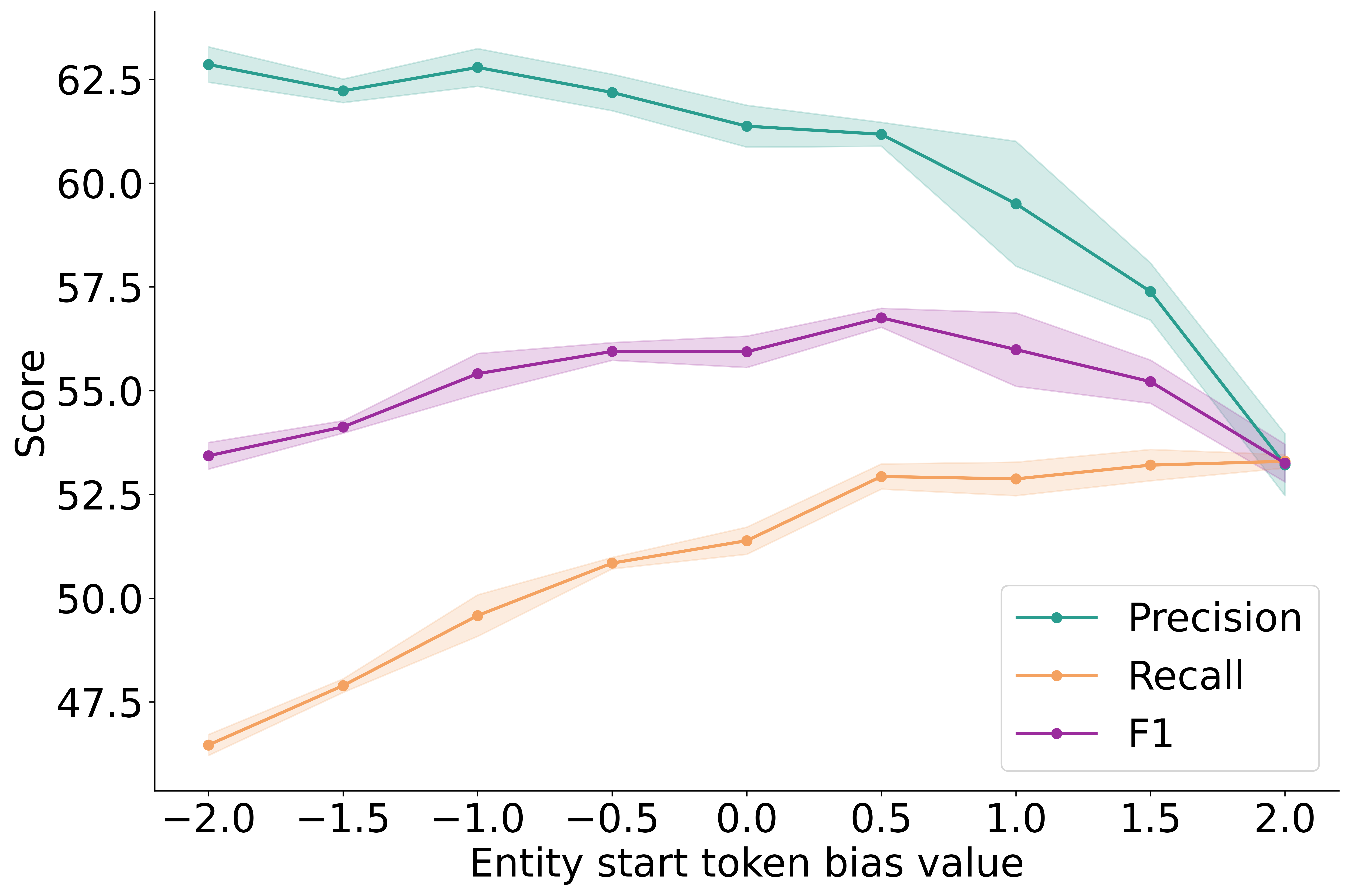}
    \caption{\textit{Entity bias value:} The effect of adding a bias to the logits of the entity start token (\texttt{<}). This simple approach allows the user significant control over the precision-recall tradeoff.}
    \label{fig:precision_recall}
\end{figure}

\subsection{Model Architecture and Training}

Denote by $\mathbf{x}$ the input audio features, which are fed into the encoder. The encoder produces a sequence of hidden states $\mathbf{h} = \text{Encoder}(\mathbf{x})$, which are then used to condition the decoding process. In our approach, we further condition the decoding process on a set of entity tags $\mathbf{t} = [t_1, t_2, \ldots, t_k]$, where each $t_i$ represents a specific entity type such as \texttt{occupation},  \texttt{celestial-body}, etc. 

The decoder generates each token $y_t$ based on the previous tokens and the encoder's hidden states, $$y_t = \text{Decoder}(y_{1:t-1}, \mathbf{h}, \mathbf{t}).$$
Overall, the output of the decoder is a sequence of tokens $\mathbf{y} = [y_1, y_2, \ldots, y_n]$, comprising both the transcribed text and the corresponding entity tags, as illustrated in  Figure~\ref{fig:arch}.

The model is trained to minimize the standard cross-entropy loss between the predicted output sequence $\mathbf{y}$ and the ground truth sequence $\mathbf{y}^*$, which includes both the correct transcription and the correct entity tags:
$$\mathcal{L}(\mathbf{y}, \mathbf{y}^*) = - \sum_{t=1}^n \log P(y_t = y_t^* \mid y_{1:t-1}, \mathbf{h}, \mathbf{t}).$$

To mark the entities in the output sequence, we explore two approaches. The first approach uses the BIO tagging scheme, similar to the approach used in \cite{ding2024rethinking}, i.e., \texttt{The(O) astronaut(B-occupation) explored(O)...}. The second marks entities with start and end markers as follow \texttt{<}$t_i$\texttt{>}$y_{i_1}...y_{i_n}$\texttt{<}$t_i$\texttt{>>...}, for example, \texttt{The <occupation>astronaut<occupation>> explored} (see Figure~\ref{fig:arch}). While the former may induce greater supervision and improve generalization~\cite{ding2024rethinking}, this approach increases the length of the generated sequence, hence it is potentially less computationally and time efficient.

Our prompting based training procedure, allows WhisperNER to handle both a predefined set of entities and open-type NER tasks. By exposing the model to diverse entity types during training, we ensure that it can generalize well to new and unseen entities during inference, making it highly adaptable to a wide range of applications.


\begin{figure}[t]
    \centering
    \includegraphics[width=1.\linewidth]{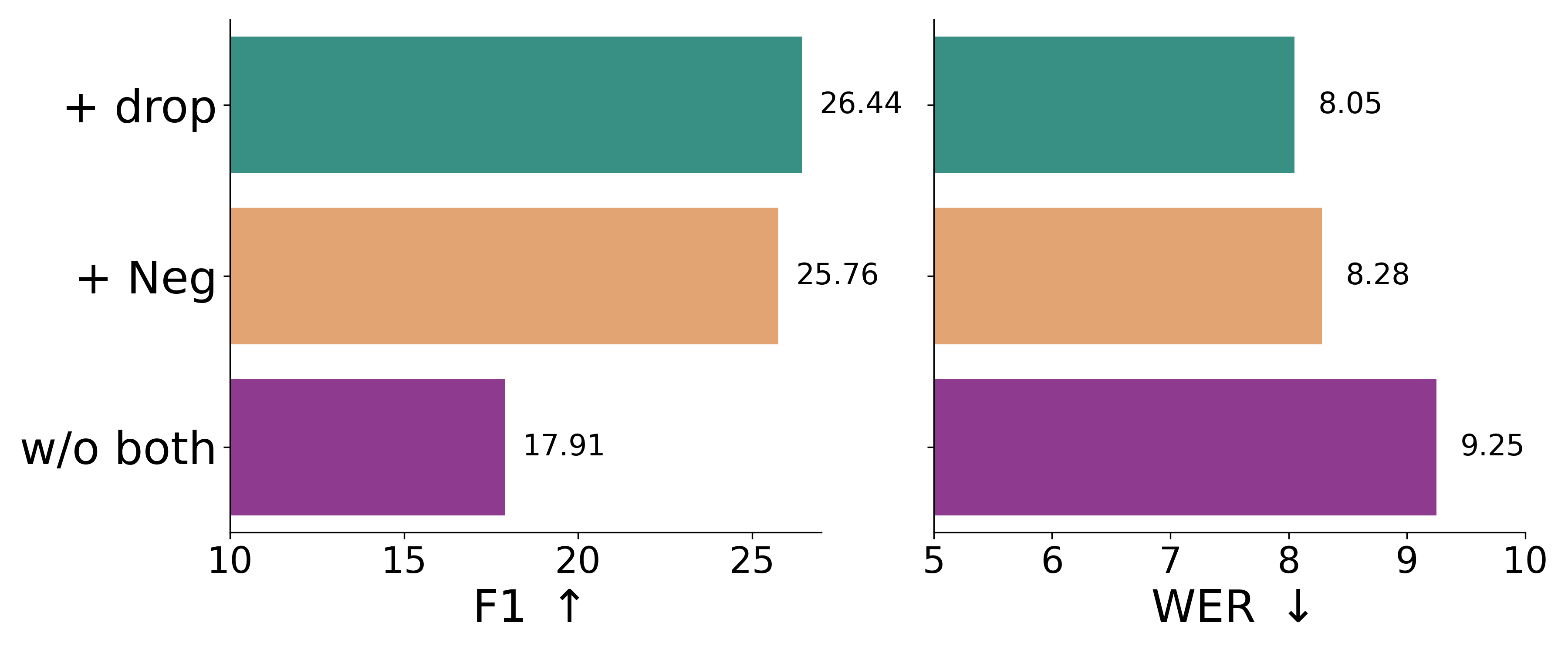}
    \caption{\textit{Negative samples and entity dropping:} The effect of incorporating negative entity tags and entity tag dropping.}
    \label{fig:neg}
\end{figure}

\subsection{Precision-Recall Control via Logit Modification}\label{sec:tradeoff}

In standard discriminative NER models, entity probabilities are explicitly computed, allowing precision-recall trade-offs to be adjusted by varying the decision threshold~\cite{zaratiana2023gliner,lou2023universal,liu2023rexuie,zhu2023mirror}. 
However, in generative models like LLMs~\cite{ashok2023promptner,wang2023gpt,xie2023self} or end-to-end approaches with a generative decoder~\cite{arora2024universlu,li2024prompting,arora2023study}, as well as in \ourmethod{}, probability-based thresholding is not directly applicable due to the model’s autoregressive nature and the implicit generation of entity spans.

To address this, we propose a precision-recall control method by adjusting the logits of the entity start token, \texttt{<}, during inference. Specifically, we introduce a tunable bias term to the raw logits (pre-softmax) of the entity start token, shifting their relative confidence levels. A positive bias enhances entity extraction, improving recall, while a negative bias makes predictions more conservative, favoring higher precision. This straightforward yet effective approach allows users to control the precision-recall tradeoff without modifying the model architecture or requiring retraining.
We evaluate this method in Section~\ref{sec:tradeoff_res} and show it allows significant control over the precision-recall tradeoff curve, which may be essential for real-world applications ~\cite{minkov2006ner,sun2022adjusting}.

\subsection{Negative Sampling and Entity Type Dropout}

Empirically, we found that prompting the model with entity labels that are absent from the speech utterance (negative entity labels) significantly improves its generalization performance. This aligns with previous findings in the literature on open-type NER~\cite{zaratiana2023gliner}. The negative set of entity labels was taken from a randomly selected example. Additionally, we applied random entity type dropout during training, as inspired by Sainz \emph{et al.} (2023)~\cite{sainz2023gollie}. This involves randomly eliminating a subset of entity types from the prompt and output sequence, compelling the model to focus on the remaining defined entity types. We found this approach to reduce entity hallucination risks during inference -- where the model outputs entities not presented in the input prompt. Also, we incorporate random shuffling of entities in the input prompt to further promote generalization across diverse input sequences. For a detailed analysis of these methodologies and their impacts, see Section~\ref{sec:ablation}.

 \begin{figure}[t]
    \centering
    \includegraphics[width=.9\linewidth]{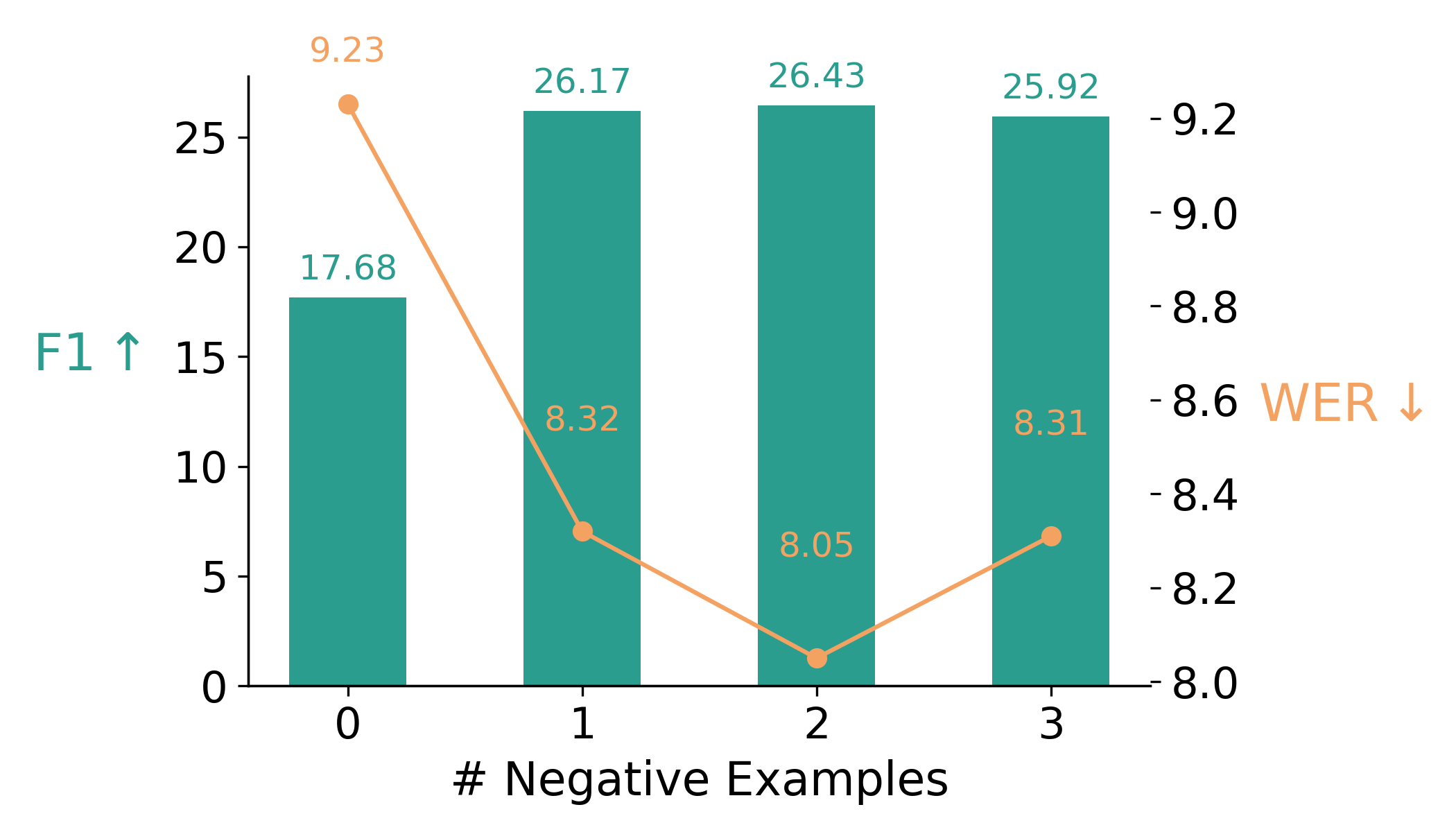}
    \caption{Effect of the number of negatives example. The x-axis describes the number of negatives examples used to construct negative NER labels.}
    \label{fig:neg_prop}
\end{figure}

\section{Experiments}
\label{sec:exp}

In this section, we evaluate \ourmethod{} against natural pipeline-based baselines using open-type NER speech datasets and common NER. To construct a dataset for evaluation, we either augment common NER text datasets with corresponding synthetic speech utterances or annotating the transcription of speech dataset with open-type NER labels. 
To support future research on joint ASR and NER, we will release all datasets.

\textbf{Datasets.} \quad 
For the open type NER benchmarks we utilize three frequently used speech datasets with diverse linguistic and contextual properties. These datasets include VoxPopuli~\cite{wang2021voxpopuli}, LibriSpeech~\cite{panayotov2015librispeech}, and Fleurs~\cite{conneau2023fleurs}, each offering unique challenges and opportunities for evaluating NER models, particularly in open-type NER contexts. For the multilingual datasets, we focus specifically on the English subset of the data. As these ASR datasets do not include NER labels, we adopt an LLM-based annotation approach, in line with the method described in~\cite{bogdanov2024nuner}.  
See~\cite{bogdanov2024nuner} for full details of the LLM-based annotation method. This annotation scheme results in the following ASR-NER datasets: (i) \textit{VoxPopuli-NER}: This dataset comprises $879$ samples. It is annotated with $2469$ unique entity types, providing a comprehensive basis for political and geographical entity recognition such as: Political Role, Demographic Group, Political Institution, Political Strategy and Social Phenomenon; (ii) \textit{LibriSpeech-NER}: Based on the LibriSpeech-clean dataset, with $1604$ samples derived from audio book readings, includes annotations for $3674$ unique entity types. This dataset is particularly valuable for testing the model’s ability to identify personal names and locations within narrative contexts; (iii) \textit{Fleurs-NER}: Consists of $441$ samples, annotated for $1440$ unique entity types. 

In addition, we utilize commonly used, textual NER benchmarks~\cite{zaratiana2023gliner,bogdanov2024nuner,sainz2023gollie,ding2024rethinking}, namely MIT-Movie~\cite{liu2013asgard} and MIT-Restaurant~\cite{liu2013asgard}. These datasets consists of $\sim14$K and $\sim10$K, with $12$ and $8$ entity types, respectively.  To enhance these text-based datasets for speech recognition, we generated corresponding audio data using the AWS Polly text-to-speech (TTS) service.

\textbf{Baselines.} \quad We evaluate our \ourmethod{} againt natural pipeline-based baselines, in which we initially employ a Whisper large-v2 model for speech-to-text transcription, followed by a NER model. We constrain our evaluation to recent small-scale NER models (up to 0.5 billion parameters) to ensure competitiveness and maintain parity in terms of the number of parameters and computational requirements, aligning with our end-to-end approach. 
Importantly, current end-to-end models support only a fixed and predefined set of entity types, making direct comparison on the open-type NER benchmarks unsuitable~\cite{arora2024universlu,li2024prompting,yang2023chinese,arora2023study}.
We evaluate the following approaches: (i) \textit{GLiNER}~\cite{zaratiana2023gliner}: a transformer encoder based architecture with zero-shot NER capabilities. We use the large variant, GLiNER-L; (ii) \textit{NuNER}~\cite{bogdanov2024nuner}: since the original NuNER model is not suitable for the zero-shot NER task, we utilize a GLiNER based architecture model, trained with the synthetic dataset of~\cite{bogdanov2024nuner}, named \texttt{NuNER zero-span}.; (iii) \textit{GNER}~\cite{ding2024rethinking}: Different from the previous two benchmarks, this approach uses generative models specifically adopted for the NER task. We use the \texttt{flan-t5-base} variant; (iv) \textit{\ourmethod{}-BIO}: Our approach with complete BIO annotations; (v) \textit{\ourmethod{}}: Our approach with NER span boundary annotations. 
(Figure~\ref{fig:arch}).

\textbf{Evaluation metrics.} \quad 
We assess the performance of our system using two evaluation metrics: Word Error Rate (WER), which quantifies the accuracy of the automatic speech recognition transcription, and F1 Score, which evaluates the effectiveness of named entity recognition. For a NER span prediction to be deemed correct, both the transcription must be accurate and the assigned NER label must match the true entity. This dual requirement ensures that our evaluation reflects both the precision of the speech recognition and the correctness of the entity identification. For the baseline methods, we compute the WER metric on the ASR's output.

\begin{table}[t]
\small
\centering
\caption{Effect of negative sampling approach: We evaluate \ourmethod{} models trained using the NuNER dataset for 25K steps, on the zero-shot task over the MIT NER Benchmarks.}
\begin{adjustbox}{max width=0.8\columnwidth}
\begin{tabular}{lcccc}
\toprule
& F1 $\uparrow$ 
& WER $\downarrow$ \\
 \midrule
 Random NER Type & $20.22$  &   $\underline{8.36}$ \\
 Hard Negatives &  $\mathbf{25.91}$  &   $8.46$ \\
 Random Sample & $\underline{25.83}$  &   $\mathbf{8.09}$ \\
 \bottomrule
\end{tabular}
\end{adjustbox}
\label{tab:ablation_neg_sampling}
\end{table}

 

\textbf{Training details.} \quad The \ourmethod{} models are trained using a subset of $350$K samples from the dataset proposed in~\cite{bogdanov2024nuner}. For each example, we generate a corresponding synthetic audio sample. Unless stated otherwise, we train the \ourmethod{} models for $250$K steps, with learning rate of $1e-6$ and linear decay learning rate scheduler. During training, we freeze the Whisper encoder and only modify the decoder.

\subsection{Zero-shot Open Type NER}

We first evaluate \ourmethod{} on the challenging open-type NER setup, in which the model is prompted at inference time with novel entity types, not observed during training.
We use the three open-type NER benchmarks, namely VoxPopuli-NER, LibriSpeech-NER, and Fleurs-NER. 
Similar to \cite{bogdanov2024nuner}, each test example is annotated with positive NER tags, while negative entities are generated by randomly sampling NER types from the full dataset.
We ensure that the number of negative NER labels matches the number of positive labels, resulting in a balanced dataset. The results are presented in Table~\ref{tab:zero_shot_open}. both \ourmethod{} variants outperform all baseline methods in terms of NER F1 score, with only a slight decrease in WER, while being more parameter-efficient. In comparison, both GLiNER-L and NuNER add 459M parameters each ($\sim$30\% increase), while GNER adds 248M parameters ($\sim$16\% increase), all on top of the Whisper large backbone.


\subsection{Controlling the Precision-Recall Tradeoff}\label{sec:tradeoff_res}

Controlling the precision-recall tradeoff is essential in some domains. For example, higher recall is crucial in medical diagnostics to avoid missing critical entities.
As described in Section~\ref{sec:tradeoff}, biasing the start entity token in the logit processor enables control over the precision-recall tradeoff in entity extraction. A negative bias enhances precision but lowers recall by suppressing entities, 
while a strong positive bias increases recall at the possible cost of false positives.
In Figure~\ref{fig:precision_recall}, we visualize our entity biasing approach using the VoxPopuli-NER dataset. As can be seen, this simple inference time modification allows significant control over the precision-recall tradeoff curve.
 

\begin{table}[t]
\small
\centering
\caption{Supervised FT Performance on the MIT NER Benchmarks.}
\begin{adjustbox}{max width=0.85\columnwidth}
\begin{tabular}{lcccccccccccccccccccccccc}
\toprule
 & \multicolumn{2}{c}{Movie} && 
 \multicolumn{2}{c}{Restaurant} 
 \\ 
 \cmidrule{2-3} \cmidrule{5-6} 
 &  F1 $\uparrow$ 
& WER $\downarrow$  &&
F1 $\uparrow$ 
& WER $\downarrow$ 
\\
 \midrule
 GNER-T5-base & $61.12$ & $3.34$ &&  $57.90$  & $5.48$ \\
 NuNER & $75.25$ & $3.34$ &&  $71.40$  & $5.48$ \\
 GLiNER & $75.34$ & $3.34$ &&  $71.09$  & $5.48$ \\
 \midrule
  \ourmethod{}-BIO  & $\underline{80.91}$ & $\underline{2.64}$ && $\mathbf{72.74}$ & $\underline{4.74}$\\
 \ourmethod{} & $\mathbf{81.35}$ & $\mathbf{2.31}$ &&  $\underline{71.51}$ & $\mathbf{4.02}$ \\
 \bottomrule
\end{tabular}
\end{adjustbox}
\label{tab:finetune_mit}
\end{table}

\subsection{Ablation}\label{sec:ablation}

To better understand the impact of negative sampling strategies and entity type dropout on model performance, we conducted an ablation study. We analyzed varying the proportion of negative samples, the method for negative sampling, and the inclusion of random entity type dropout during training. For efficiency and consistency, all models were trained for 25K steps using the augmented NuNER dataset~\cite{bogdanov2024nuner} and evaluated on the MIT NER benchmarks.
Table~\ref{tab:ablation_neg_sampling} shows that using full entity sets from negative examples, ``random sample'', consistently outperforms random negative sampling of NER types, likely due to the contextual consistency of the selected examples. We also tested a ``hard negative'' approach, where negative examples are chosen based on semantic similarity to positive examples. While this method performed comparably, it introduced additional overhead in querying the dataset, hence we opt for random sampling from the training set for efficiency.
Figure~\ref{fig:neg_prop} demonstrates that using two negative samples (approximately $66\%$ negative entity types) provides the best F1 and WER results. Figure~\ref{fig:neg} highlights the effectiveness of random entity type dropout, which helps reduce entity tag hallucinations during inference. Based on these findings, we train \ourmethod{} with entity type dropout and negative sampling based on the entity set from two negative examples.

\begin{figure}[t]
    \centering
    \includegraphics[width=1.\linewidth]{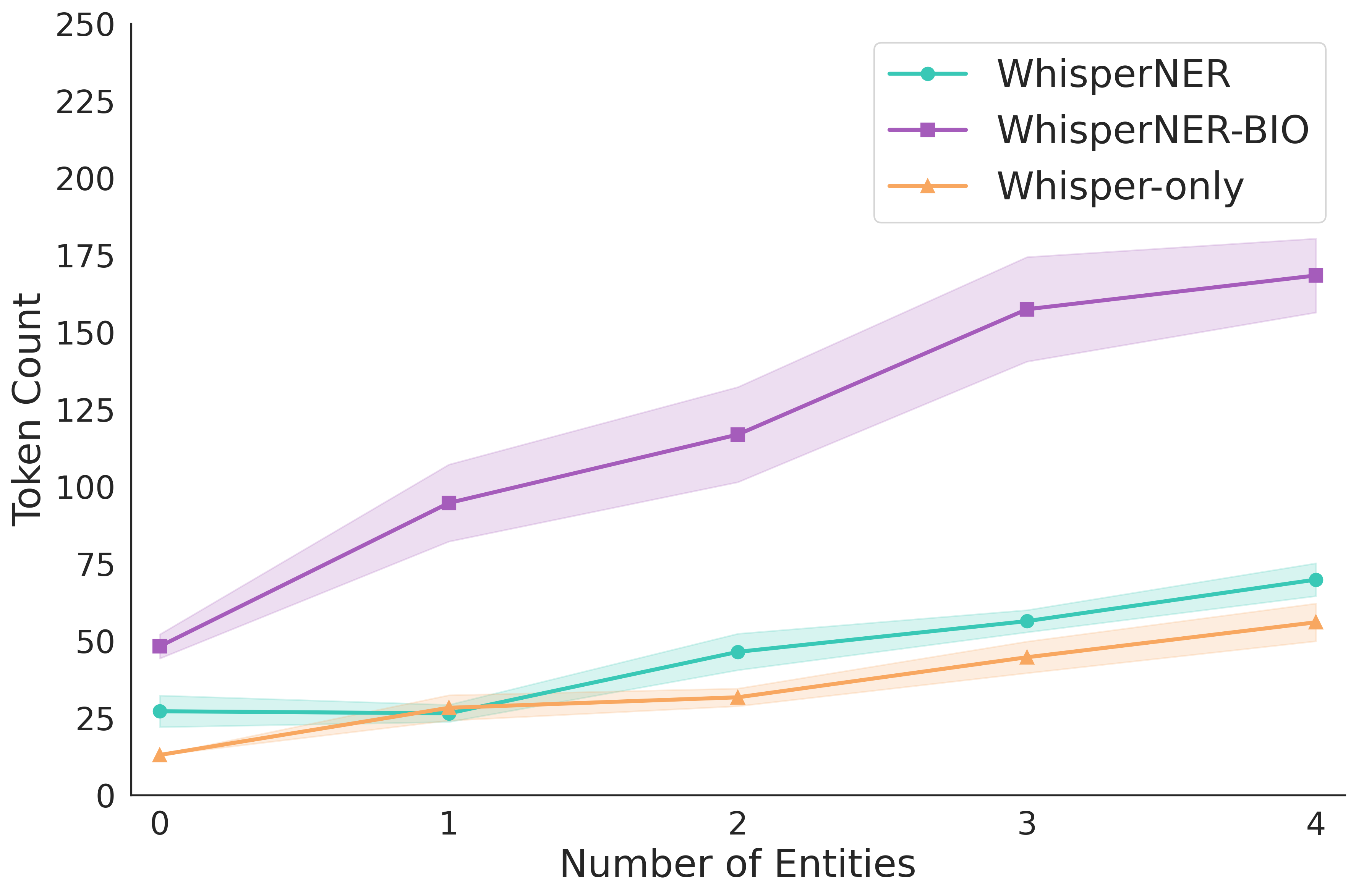}
    \caption{\textit{Sequence Length Analysis:} The BIO tagging scheme significantly increases token count, while the WhisperNER format keeps predicted sequence length much closer to the Whisper-only baseline.}
    \label{fig:sequence_length_analysis}
\end{figure}

\subsection{Supervised Finetuning}
While the main focus of our paper is on the challenging open-type NER benchmark, some scenarios requires operating on a fixed set of entities. Here, we evaluate the supervised fine-tuning performance of our WhisperNER model using commonly used closed set NER dataset from the MIT NER benchmark~\cite{liu2013asgard}, namely MIT-Movie and MIT-Restaurant. For baseline methods, we independently finetune both the Whisper model and the NER model.
The results are summarized in Table~\ref{tab:finetune_mit}. 
Our method outperforms all baselines, achieving the best performance in both WER and NER F1 scores. 
These results highlight \ourmethod{}'s effectiveness in handling transcription and entity recognition 
in a supervised setting.

\subsection{Runtime Analysis}

We analyze the predicted sequence length as a proxy for runtime and computational efficiency. Specifically, we compare WhisperNER (start/end marker format), WhisperNER-BIO (BIO tagging scheme), and the original Whisper-only baseline. Our findings presented in Figure~\ref{fig:sequence_length_analysis} show that the BIO tagging scheme introduces a significantly higher number of tokens, increasing the overall output length and, consequently, the decoding cost. In contrast, the start/end marker format used in Whisper-NER adds substantially fewer special tokens, resulting in predicted sequences that are much closer in length to the Whisper-only baseline. This suggests that Whisper-NER is more efficient at inference time than its BIO variant, enabling cost-effective joint transcription and entity recognition.

\subsection{Zero-shot Language Generalization}

To assess the cross-lingual capabilities of WhisperNER, we conduct a preliminary zero-shot evaluation on languages not seen during training. We apply the English-trained model to open-type NER tasks in Spanish and French using the VoxPopuli dataset. As shown in Table~\ref{tab:zeroshot_lang}, the model exhibits some generalization ability, achieving F1 scores of 29.3 and 28.6 on Spanish and French, respectively, with relatively low word error rates. These results suggest that our approach can transfer to other languages to some extent, even without exposure during training. This points to the potential of extending WhisperNER to new languages through multilingual pretraining or lightweight fine-tuning on small labeled datasets.

\begin{table}[t]
\centering
\caption{Zero-shot generalization to unseen languages using the English-trained WhisperNER model. Evaluated on VoxPopuli.}
\label{tab:zeroshot_lang}
\begin{tabular}{lcc}
\toprule
Language & F1 $\uparrow$ & WER $\downarrow$ \\
\midrule
Spanish & 29.3 & 4.8 \\
French  & 28.6 & 13.9 \\
\bottomrule
\end{tabular}
\end{table}

\section{Limitations}

While \ourmethod{} demonstrates strong performance in end-to-end transcription and NER, several limitations remain. First, as described in Section II, we adopt an LLM-based annotation approach to generate NER labels for ASR datasets that lack human annotations, following the method proposed in~\cite{bogdanov2024nuner}. While this enables scalable and cost-effective supervision, relying on an LLM for both training and test annotations may introduce systematic biases and reduce the independence of the evaluation set. As a result, model performance may reflect alignment with the LLM’s labeling behavior rather than genuine generalization from the speech model.
Second, we observe a modest degradation in word error rate, which we hypothesize is due to fine-tuning on synthetic speech generated via text-to-speech. Although TTS supports scalable data augmentation, it may introduce mismatches in acoustic characteristics when compared to natural speech.
Third, our model introduces additional generation time due to the inclusion of NER tag tokens in the output sequence, which increases decoding length. While this overhead is inherent to the joint modeling strategy, future work may explore more efficient decoding or selective tagging approaches to reduce computational cost.

\section{Conclusion}

In this paper, we propose WhisperNER, a novel approach for joint open type named entity recognition and automatic speech recognition. Our model is designed to generalize to novel entities at inference, effectively handling a diverse range of entity types. We demonstrate that WhisperNER not only outperforms traditional pipeline-based baselines in terms of entity recognition accuracy but also achieves this with minimal to no increase in transcription errors. Our work marks a step forward in the integration of speech and NLP, significantly improving both the effectiveness and precision of speech applications.

\bibliographystyle{IEEEtran}
\bibliography{ref}

\end{document}